\documentclass[%
 reprint,
superscriptaddress,
 amsmath,amssymb,
pre,
]{revtex4-1}

\usepackage{graphicx}
\usepackage{dcolumn}
\usepackage{bm}

\begin{document}

\preprint{APS/123-QED}
\bibliographystyle{apsrev4-1.bst}

\title{Computing Motion with 3D Memristive Grids}

\author{Chuan Kai Kenneth Lim}%
 \email{chuan.lim08@imperial.ac.uk}
\affiliation{%
 Centre for Bio-inspired Technology, Department of Electrical and Electronic Engineering,Imperial College London, London SW7 2AZ, United Kingdom
}%
\author{T.Prodromakis}
\email{t.prodromakis@soton.ac.uk}
\affiliation{%
 Centre for Bio-inspired Technology, Department of Electrical and Electronic Engineering,Imperial College London, London SW7 2AZ, United Kingdom
}%
 \affiliation{School of Electronics and Computer Science, University of Southampton, Southampton SO17 1BJ, United Kingdom}

\date{\today}

\begin{abstract}
Computing the relative motion of objects is an important navigation task that we routinely perform by relying on inherently unreliable biological cells in the retina. The non-linear and adaptive response of memristive devices make them excellent building blocks for realizing complex synaptic-like architectures that are common in the human retina. Here, we introduce a novel memristive thresholding scheme that facilitates the detection of moving edges. In addition, a double-layered 3-D memristive network is employed for modeling the motion computations that take place in both the Outer Plexiform Layer (OPL) and Inner Plexiform Layer (IPL) that enables the detection of on-center and off-center transient responses. Applying the transient detection results, it is shown that it is possible to generate an estimation of the speed and direction a moving object.
\end{abstract}

\pacs{Valid PACS appear here}
\maketitle


\section{\label{sec:level1}Introduction}

Computing motion is essential for performing many daily tasks that also find applications in machine vision\cite{RefWorks:145,RefWorks:143}, industrial automation \cite{RefWorks:149,RefWorks:148,RefWorks:146,RefWorks:147} and robotics\cite{RefWorks:142,RefWorks:144}. Traffic engineers and transport authorities require motion computation and object tracking in order to implement effective intelligent transportation systems\cite{RefWorks:150,RefWorks:151}. For the defense industry, the ability to track mobile enemy targets\cite{RefWorks:152} and control unmanned vehicles\cite{RefWorks:153} is critical. The tracking of cell movements is also recently shown to play an important role in medical diagnostic procedures \cite{RefWorks:154,RefWorks:155}. While in the field of communications, combining motion estimation and tracking with predictive coding can allow for significant reduction in transmission bandwidth \cite{RefWorks:30}. 

By utilizing parallel networks and hierarchical structures, biological systems are able to perform motion computation very efficiently\cite{RefWorks:158}. In comparison, conventional motion computation systems typically utilize charged-coupled device (CCD) cameras and digital processors that result in time-sequential computations. As a result, large processing power is required to ensure that these serial computations can be performed in real time. In recent years, the need for physically smaller and more power efficient devices has driven research into neuromorphic\cite{RefWorks:159,RefWorks:160}and biomimetic systems\cite{RefWorks:161}, which are largely inspired by the parallelization naturally found in biological systems.

One of the first biomimetic motion computation systems was implemented by a team lead by Carver Mead in Caltech\cite{RefWorks:127,RefWorks:121}. By modeling the optical flow via a network of analog resistive devices, this approach was shown to be able to compute an estimate of the optical flow. Over the past two decades, various other biomimetic approaches to motion computation have been implemented in CMOS technology\cite{RefWorks:127,RefWorks:111,RefWorks:130,RefWorks:35,RefWorks:7,RefWorks:49,RefWorks:47,RefWorks:56,RefWorks:39,RefWorks:44,RefWorks:46,RefWorks:40,RefWorks:42,RefWorks:41}. 

Perhaps, the most prominent implementation is a 128x128 asynchronous temporal contrast silicon retina developed by Delbruck\cite{RefWorks:111}. This system makes use of address-event representation (AER)\cite{RefWorks:162}; inspired by the way neurons utilise spiking events for communication. Essentially, the AER approach allows the transmission of only the local changes caused by movement in a scene, instead of transmitting at a fixed frame rate. This increases the computational efficiency of the chip, enabling it to perform over an impressive dynamic range of more than 120 dB with a low power consumption of 23mW. 

Another notable implementation is Visio1\cite{RefWorks:130}, a retinomorphic chip with parallel pathways that mimics the OPL and IPL. Subthreshold current-mode circuits are used to model the autofeedback characteristics of horizontal cells (for spatial filtering) and the loop-gain modulation of amacrine cells (for adapting temporal filtering to motion). Detection of edges moving in one direction or the other is made possible by aVLSI implementation of ganglion cells that respond to motion in a quadrature sequence. 

The recent discovery of practical memristive devices \cite{RefWorks:137}  has opened the option of improving on these CMOS motion sensors. It has been shown in \cite{RefWorks:23} that memristor-MOS technology (MMOST) is capable of outperforming CMOS implementations in terms of power and size. Furthermore, the non-linear and adaptive response of memristors allow them to serve as excellent building blocks for the practical realisation of complex synaptic connections\cite{RefWorks:27,RefWorks:163}. Our group has previously demonstrated that memristive grids have a great potential for modeling certain aspects of early vision processing, which takes place in the OPL\cite{RefWorks:19}. 

Here, we describe a novel approach, distinct from the current implementations \cite{RefWorks:127,RefWorks:111,RefWorks:130,RefWorks:35,RefWorks:7,RefWorks:49,RefWorks:47,RefWorks:56,RefWorks:39,RefWorks:44,RefWorks:46,RefWorks:40,RefWorks:42,RefWorks:41} 
, that allows for motion computation based on memristive networks, by extending our results in \cite{RefWorks:19}. A novel memristive thresholding scheme is utilized for enabling the tracking of moving objects within a scene.  Also, a biomimetic model of both the OPL and IPL are described that facilitates the detection of ON and OFF center transient responses. In turn, we investigate the directional and speed computation capacity of our approach. Finally, we demonstrate that this platform can operate in a fault tolerant manner and can adapt to distinct lighting conditions, similarly to biological counterparts. 

\section{\label{sec:level1}Background}

\subsection{\label{sec:level2}Vertebrate's Retina: Architecture and Cells}
In order to perform any form of biomimicry, it is necessary to comprehend the architectural and functioning features of the leveraging biological system. This section serves as a brief unqualified introduction of the retina and is not meant to be a thorough review. More in depth information can be found in  \cite{RefWorks:82,RefWorks:135,RefWorks:136}. A simplified drawing showing the different layers of the human retina can be found in Fig 3 of \cite{RefWorks:3} and Fig 1of \cite{RefWorks:19}.

There are 5 main types of retina cells (photoreceptor cells, bipolar cells, horizontal cells, amacrine cells and ganglion cells) laid along two major layers: the Outer Plexiform layer (OPL) and the Inner Plexiform layer (IPL). The photoreceptors (rods and cones) receive visual stimuli from the outside world. The outputs of the photoreceptor cells form synaptic connections with bipolar cells and horizontal cells. The horizontal cells are interneurons that form lateral local connections. The bipolar cells connect the OPL, comprising of horizontal and photoreceptor cells, to the IPL containing the amacrine cells and ganglion cells. Similar to horizontal cells, amacrine cells mediate synaptic lateral interactions between bipolar cells and ganglion cells in the IPL.

The photoreceptor cells essentially act as transducing elements, transforming the incident light stimuli into electrical signals. The magnitude of voltage change in the cells membrane is proportional to the logarithm of the intensity of light\cite{RefWorks:39}. Horizontal cells form synaptic connections between photoreceptors and bipolar cells. Via the horizontal cells, lateral connections with neighbouring groups of photoreceptors and bipolar cells are made. These can be conceived as being a resistive layer spanning the OPL, with the architecture allowing dynamic range adjustments. 

The output of a bipolar cell depends on the difference between the voltage at the photoreceptor it is connected to and the voltage at neighbouring photoreceptors, through the indirect connection via the horizontal cells. There are 2 kinds of bipolar cells. The ON-cells that respond to the onset of light and the OFF-cells that respond to the cessation of light. The photoreceptor, horizontal and bipolar cells do not generate action potential, instead, computation is performed via the flow of depolarization potentials. On the contrary, amacrine and ganglion cells in the IPL do generate spikes. Amacrine cells give transient light responses to either ON stimulus, OFF stimulus or both. Amacrine cells can be classified into sustained and transient cells. Sustained amacrine cells obtain inputs from the bipolar cells that are not inhibited by other amacrine cells. The response of sustained amacrine cells then inhibits bipolar terminals in the narrow field region, causing those terminals to respond transiently. Essentially, this network of amacrine cells forms a high pass filter of the bipolar signal that facilitates the detection of moving edges\cite{RefWorks:43}.

Ganglion cells carry the output of the retina to the optical nerve, which leads to the visual cortex in the brain. Of the different types of ganglion cells, the directionally selective (DS) ganglion cells play a huge role in motion detection. There are three identified types of DS ganglion cells, namely ON/OFF DS ganglion cells, ON DS ganglion cells and OFF DS ganglion cells. ON/OFF DS ganglion cells perform the function of local motion detectors.

\subsection{\label{sec:level2}Mathematical Model of Memristors}
The derivation of the mathematical memristor model employed here is largely guided by \cite{RefWorks:97}. A memristive system was defined by Chua and Kang\cite{RefWorks:110}, following the original 1971 definition of the ideal memristor\cite{Chua1971}, as:
\begin{equation}
 v = R(x)i
\end{equation}
\begin{equation}
 \frac{dx}{dt} = f(x,i)
\end{equation}
This implies that the resistance $R(x)$ depends on an internal state $x$ of the device while the time derivative of the internal state $x$ is a function of $x$ and $i$. The first physical model was conceived of a solid-state implementation by Strukov \textit{et al.}\cite{RefWorks:96}, where $x$ is proportional to charge $q$ flowing through it. In the device, the magnitude of $R$ can be modified reversibly between a highly conductive state $R_{on}$ and a highly resistive state $R_{off}$, by modulating $x$:
\begin{equation}
R(x)= x(t) R_{on}+(1-x(t))R_{off}
\end{equation}
$x(t)$ is restricted to the interval [0,1] and the time derivative of $x(t)$ is proportional to the current, as shown:
\begin{equation}
 \frac{dx}{dt} = \frac{R_{on}}{\beta } i(t) \
\end{equation}
Substituting (3) and (4) into (1) we obtain:
\begin{equation}
v(t)= \beta \left\{ x(t)+r[1-x(t)]\right\}\frac{dx(t)}{dt}
\end{equation}
where $r = R_{off}/R_{on}$ is the resistance ratio and $\beta$ has a dimension of magnetic flux. Moreover, since $\varphi =\int v dt$, and by using $x \frac{dx}{dt} =  \frac{1}{2} \frac{d}{dt} x^2$, it is possible to integrate both sides of (5) to obtain:
\begin{equation}
\varphi = \beta [- \frac{r-1}{2}x^2 + rx+c]
\end{equation}
where $c$ is a constant of integration determined by the initial conditions of $x$. This equation shows that flux is a quadratic function of the charge as $x$ is proportional to $q$, hence, the non-linear relationship\cite{RefWorks:137,salaoru2013}.

Solid-state $TiO_2$ memristors implementations typically comprise two metal contacts that encompass a $TiO_2$ active core of thickness $D$. An external bias across the device causes charge to flow through the device. This causes a drifting of dopants resulting in the movement of the boundary between the two regions. By defining $w(t)$ as the coordinates of this boundary and $\mu_v$ as the average ion mobility, equation (4) becomes:
\begin{equation}
\frac{1}{D}\frac{dw(t)}{dt} = \mu_v\frac{R_{on}}{D^2} i(t)
\end{equation}
and by integrating both sides, we obtain:
\begin{equation}
\frac{dw(t)}{D} = \mu_v\frac{R_{on}}{D^2} q(t)
\end{equation}
It is observed that $x(t) = w(t)/D$ corresponds to the normalized width of the $R_{on}$ region while $\beta = D^{2}/\mu_v$. Setting $c$ to zero in equation (6) and substituting equation (8):
\begin{equation}
\varphi = - \frac{R_{on}\mu_v}{2D^2} (\frac{R_{on}}{R_{off}} - 1)q^2 + R_{off}q
\end{equation}
Assuming that $R_{off}\texttt{>>}R_{on}$, the memristance of the device is obtained as:
\begin{equation}
M(q) = \frac{d\varphi}{dt} = R_{off}(1 - \frac{R_{on}\mu_v}{D^2}q)
\end{equation}
Furthermore, as mentioned in \cite{RefWorks:96}, there are significant non-linearities in ionic transport, especially in the thin film edges at the boundary, which can be modeled by a window function,$f(x)$, on the right hand side of equation (7), giving:
\begin{equation}
\frac{dx(t)}{dt} = ki(t)f(x), k = \mu_v\frac{R_{on}}{D^2}, x(t) = \frac{w(t)}{D}
\end{equation}

\subsection{\label{sec:level2}Optical Flow}
The estimation of the direction and speed of a pixel in a sequence of image can be obtained from the optical flow; the approximation of local motion in an image by computing local spatial and temporal derivatives in a given sequence of frames. For a 2D picture, the optical flow indicates the speed and direction in which each pixel in an image moves between frames. The computation of optical flow is based on spatio-temporal intensity variations in brightness patterns. In this work we assume that all intensity variations are due to the motion of an object. In general, the optical flow and the velocity of an object are different. As explained in \cite{RefWorks:83}, a perfectly featureless rotating sphere has a non-zero velocity component, but it will not give rise to any optical flow. Whereas, a shadow moving across the same sphere, which is now stationary, will produce an optical flow that is non-zero, although its velocity is zero. Nonetheless, apart from these situations, the computed optical flow is a good indication of the velocity of the moving object, provided a strong enough gradient between the moving object and its background exists. This is generally true for most natural scenes and it is assumed to be true for all input stimuli employed in this work. 

\subsubsection{2D Motion Constrain Equation}
Assuming $I(x,y,t)$ is the intensity of a brightness pattern at specific pixel located at coordinates $(x,y)$ in the image at time $t$, and if this brightness pattern moves by $(\delta x, \delta y)$ within a timeframe $\delta t$, the intensity at the corresponding pixel will be $I(x+ \delta x, y + \delta y, t + \delta t)$. Since this is essentially the same brightness pattern, it can be modeled as:
\begin{equation}
I(x,y,t)= I(x+ \delta x,y + \delta y,t + \delta t) 
\end{equation}
And by employing a Taylor expansion about the point $(x,y,t)$, this reads:  
\begin{eqnarray}
I(x+ \delta x,y + \delta y,t + \delta t) = I(x,y,t)+ \nonumber\\
  \frac{\partial I}{\partial x}\delta x+  \frac{\partial I}{\partial y}\delta y+  \frac{\partial I}{\partial t}\delta t +H.O.T
\end{eqnarray}
By comparing equation (12) and (13), it can be derived that 
$\frac{\partial I}{\partial x}\delta x+ \frac{\partial I}{\partial y}\delta y+  \frac{\partial I}{\partial t}\delta t = 0$ and $\frac{\partial I}{\partial x}\frac{\delta I}{\delta t}+ \frac{\partial I}{\partial y}\frac{\delta y}{\delta t}+  \frac{\partial I}{\partial t}\frac{\delta t}{\delta t}= 0$ and eventually,
\begin{equation}
\frac{\partial I}{\partial x}v+  \frac{\partial I}{\partial y}u+  \frac{\partial I}{\partial t}= 0
\end{equation}
This can be written more compactly as
\begin{equation}
(I_x  ,I_y )\cdot (v ,u)= - I_t = \nabla I \cdot \overrightarrow{v}
\end{equation}
$\nabla I$ refers to the spatial intensity gradient and $\overrightarrow{v}$ refers to the optical flow at pixel $(x,y)$ at time $t$.

Equation (15) has 2 unknowns, which as explained in \cite{RefWorks:83,RefWorks:34} stem from the aperture problem. Insufficient information is available to measure the full image velocity, and only the component perpendicular to the edge or along the spatial gradient can be measured. In other words, a point in an image sequence only provides one independent image measurement whereas the velocity field has two components, forming an ill-posed problem.

\subsubsection{Smoothness assumption}
In order to solve this ill-posed problem, a second constrain is therefore required. Horn and Schunck \cite{RefWorks:83} combined the 2D motion constrain together with a global smoothness term, such the final velocity field $\overrightarrow{v}$ is one which minimizes:
\begin{eqnarray}
E(u,v) = \int\int (I_x+I_y+I_t)^2 + \nonumber\\
\lambda[(\frac{\partial u}{\partial x})^2+(\frac{\partial u}{\partial y})^2+(\frac{\partial v}{\partial x})^2+(\frac{\partial v}{\partial y})^2]dxdy
\end{eqnarray}

The first term implies that the solution should be as close as possible to the measured data while the second term imposes a smoothness constrain on the solution. The smoothness constrain derives from the fact that, apart from at specific discontinuities, adjacent points on an object have similar velocity and the brightness pattern varies smoothly almost everywhere. The magnitude of $\lambda$ determines the importance of minimizing the smoothness term. With high Signal to Noise Ratio (SNR), the importance of the first term increases and $\lambda$ will be small. In contrast, if the data is unreliable, $\lambda$ will be larger, emphasizing the importance of the smoothness term.

Equation (16) shows that $E(u,v)$ is quadratic in the unknowns $u$ and $v$. From standard variation calculus it can be shown that the corresponding Euler-Lagrange equations are linear in $u$ and $v$:
\begin{eqnarray}
I_{x}^2u + I_{x}I_{y}v - \lambda\nabla^2u + I_{x}I_{t} = 0 \nonumber\\
I_{x}I_{y}u + I_{y}^2v - \lambda\nabla^2v + I_{y}I_{t} = 0
\end{eqnarray}
This results into two linear equations describing every point, which essentially capture the necessary components for calculating optical flow.

\section{\label{sec:level1}Detecting Moving Edges With Memristive Grids}
We have previously presented (in Figure 4 of \cite{RefWorks:19}) a retinomorphic OPL, which modeled the synaptic interconnections between horizontal, bipolar and photoreceptor cells. Any light stimulus is translated into an appropriate voltage source to bias the underlying hexagonal memristive network via a series resistor. It was shown that the network was able to perform a local Gaussian filtering function similar to the biological counterpart  [21], as well as edge detection via a memristor-based thresholding scheme. Nodes corresponding to edge pixels have large potential difference between neighbouring nodes due to the overlying intensity contrast at these pixels. From equation (10), the change in the memristance of the memristive fuse is proportional to the current flowing from that edge node to a neighbouring node. Therefore, by monitoring the state variance of memristive fuses associated with a node, edge detection can be performed.  Here, we extend the results of \cite{RefWorks:19} by performing detection of edges of moving objects from a video, investigating how a memristive network can be used to detect moving edges from a sequence of frames. 

\subsection{\label{sec:level2}Simulation Methods}
The employed video input stimulus is a 80x70 pixel, 0.6s segment of the video ‘traffic.mju’, which can be found in the Matlab Image Processing Toolbox\cite{Matlab_imageprocessing}. At a frame rate of 15fps, this corresponds to 9 frames, shown as a montage in Figure 1a. Changes only occur at interval of the frame period(1/15s), and within this period the voltage stimulus at each pixel is a constant. This is modeled with a piecewise-linear voltage source at each node corresponding to a pixel.

\begin{figure*}
\includegraphics[scale = 0.6]{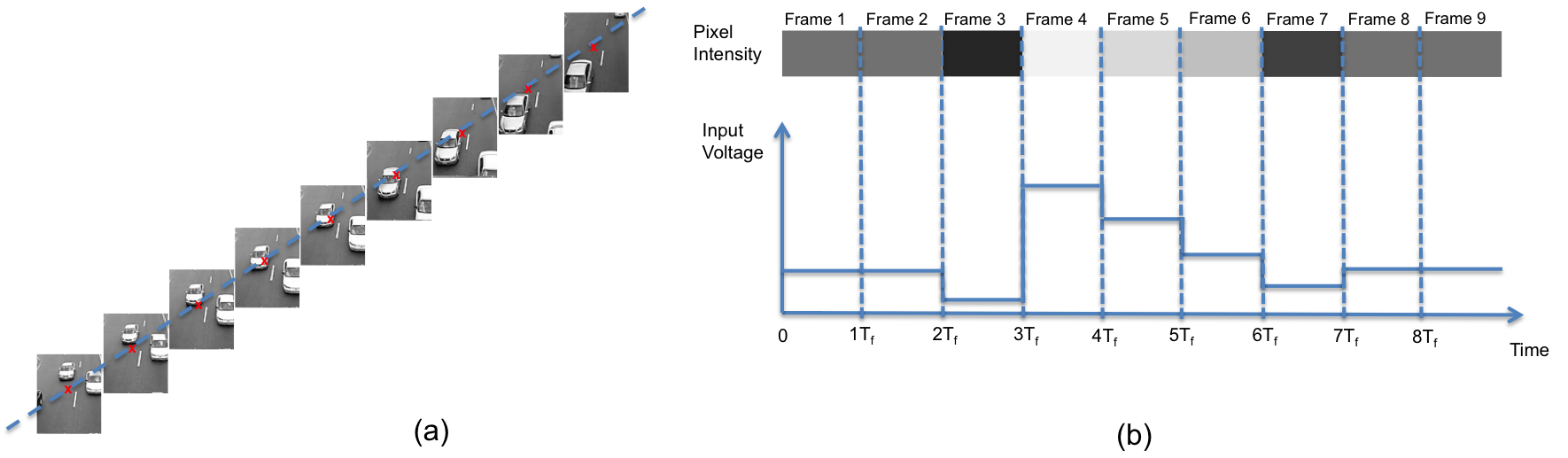}
\DeclareGraphicsExtensions{.png}
\caption{\label{fig:wide} (a) The sequence of the input frames is shown. A specific pixel in the video, marked by a Red X is indicated in each of the frames. The time between consecutive frames, $T_f$, is the inverse of the framerate. (b) The pixel intensity of each frame is shown and this corresponds to an input voltage stimulus, which is applied to the respective node in the memristive grid.}
\end{figure*}

Figure 1b describes how a single pixel in the video is converted into a voltage stimulus. In each gray scale frame, the intensity of each pixel is uniformly quantized into 256 levels, where 0 represents the absolute black and 255 represents white. Each pixel is therefore represented by an integer from 0-255. Since the maximum voltage input into each of the voltage input node in the memristive grid is set to 40mV, if the quantised intensity of a pixel is represented by X, the stimulus to its respective voltage input node is X*40/255 mV. This operation is performed for each of the pixel in each of the consecutive frames. The input voltage stimulus into a specific node corresponding to a pixel is then set to the piece-wise-linear combination of this quantised voltage level. 

MATLAB was employed to implement the netlist for the hexagonal memristive network and to convert the input frames into equivalent biasing voltage stimulus. The generated netlist is then imported into HSPICE to perform the necessary circuit simulations on the constructed system. Memristive devices were simulated with Biolek's model \cite{RefWorks:95}, with the following parameter settings: $R_{on}=100\Omega$   $R_{off}=16k\Omega$   $R_{init}=200\Omega$. In addition, the Prodromakis's windowing function \cite{RefWorks:76} was employed to take into account the non-linear dopant kinetics of the memristive elements. The results from the HSPICE simulations were then imported back to MATLAB for further analysis.

\subsection{\label{sec:level2}Detection of moving edges via memristive state thresholding}
The circuit was simulated for the input stimulus shown in Figure 1a. Edge detection was performed for each time step of the simulation, using the algorithm described in \cite{RefWorks:19}. The results are shown in Figure 2b where the edges are marked by a white pixel. 
\begin{figure}[h]
\includegraphics[scale = 0.6]{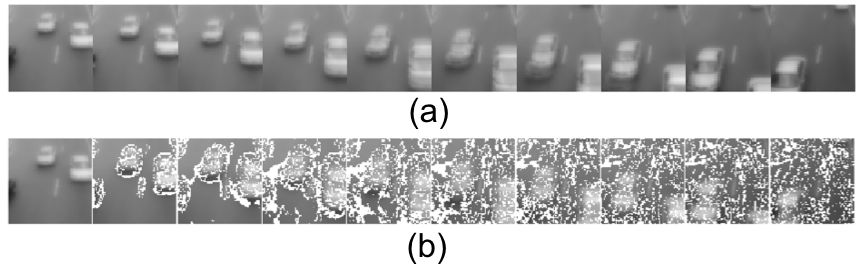}
\DeclareGraphicsExtensions{.png}
\caption{(a) Evolution of the smoothed input stimulus. (b) Simulation of a sequence of images with the edge detection algorithm from \cite{RefWorks:19}. Notice that edges were detected successfully only for the first 2 frames.}
\end{figure}
We notice that the memristance based thresholding scheme described in  \cite{RefWorks:19} was able to pick up the edges in the first 2 frames. However, for the rest of the frames, it seems like that edges were being picked up randomly. This is caused by the fact that the memristors used for the simulations are non-volatile and they do not reset or change before the next input frame is being simulated. Hence, even when the next frame is being simulated, the past memristance changes are still present, causing the results to be corrupted. In the following section, we address this issue by incorporating a novel memristive thresholding scheme that is applicable for detecting moving edges. A video showing the time evolution of the simulation is provided in S1(see Supplementary Material\cite{Supplemental}).

\subsection{\label{sec:level2}Edge detection based on monitoring the memristance modulation rate}
To overcome the problem presented presented previously, we need to be able to separate the change in memristance corresponding to moving edges in the next frame from the changes in memristance corresponding to edges in the current frame. The proposed solution exploits the tracking of the temporal derivative of the memristance of the six memristive fuses connected to each node. 

Following Equation (10), the change in memristance is proportional to the current flowing through the device, which is dependent on the potential difference across the device. Since each memristive fuse links adjacent nodes, the potential difference across it will be proportional to the intensity difference between the two nodes. If a node is an edge, there will be a strong intensity gradient with its neighbouring nodes. This results in a large rate of change of memristance for the associated memristive fuses. Hence, by monitoring this rate of change, we are able to determine if a node corresponds to an edge pixel. 

In a sequence of moving images, once the edge of the object moves away from a node, that particular node will cease to have a strong intensity gradient with its neighbouring nodes. Consequently, the temporal derivative of the memristance of the associated memristive nodes will no longer be high and hence the node will not be denoted as an edge. Figure 3 presents results that demonstrate our approach. A node is denoted as an edge pixel if the rate of memristance change of two or more memristive fuse associated with that node exceeds a predetermined threshold. Each edge pixel is denoted by a black pixel, superimposed on a white background. It is observed that the moving edges can now be successfully detected for the entire sequence of images. The full transient response of this experiment can be found in S2(see Supplementary Material\cite{Supplemental}).
\begin{figure}[h]
\includegraphics[scale = 0.6]{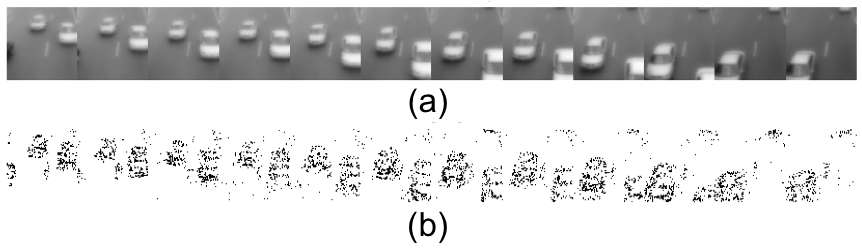}
\DeclareGraphicsExtensions{.png}
\caption{(a) Evolution of the smoothed input stimulus. (b) Moving edge detection by tracking the temporal derivative of the memristance of the fuses associated to each node.}
\end{figure}

\subsection{\label{sec:level2}Adaptation to varying lighting conditions}
Similar to \cite{RefWorks:19}, we evaluated the proposed algorithm against distinct luminance levels. Figure 4 demonstrates that even when the brightness level was halved, the moving edges were still detected successfully. The moving edge detection results under the two different lighting conditions were compared pixel by pixel and the average mismatch between these two cases was found to be 4.22\%. This is illustrated in Figure 4(c), where each white pixel corresponds to a mismatch. The full transient response of this experiment can be found in S3(see Supplementary Material\cite{Supplemental}).
\begin{figure}[h]
\includegraphics[scale = 0.6]{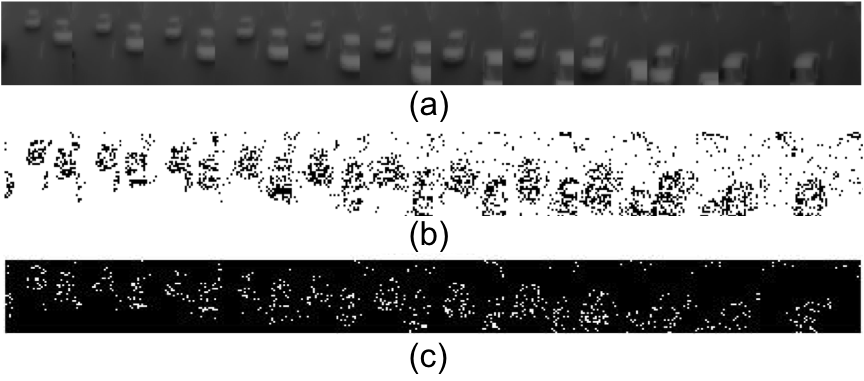}
\DeclareGraphicsExtensions{.png}
\caption{(a) Evolution of the smoothed input stimulus. (b) Moving edge detection with input stimulus of half the original brightness. (c) Illustrates the accentuated mismatch between moving edge detection under normal lighting condition and under half the original brightness.}
\end{figure}

\subsection{\label{sec:level2}Memristance Variability Tolerance}
The memristive network was tested to determine how tolerant it was to potential defects in the network. To simulate this, defect memristors were artificially introduced. Instead of simulating a circuit with all uniformly initialised memristors (i.e. $R_{on}=100\Omega$   $R_{off}=16k\Omega$   $R_{init}=200\Omega$), we now initialise a percentage of the memristors randomly. These “defected” memristors will be initialised randomly with $R_{on}=50\Omega-100\Omega$ in increments of $10\Omega$, $R_{off}=10k\Omega-20k\Omega$ in increments of $0.5k\Omega$, $R_{init}$ as a multiple (2-20) of the new $R_{on}$. The simulation results for 50\% defects and 80\% defects are respectively shown in Figure 5b and 5c. 

The memristive network appears to be rather robust to faults, even when 80\% of the devices were initialised randomly. The full transient response of this experiment can be found in S4(see Supplementary Material\cite{Supplemental}). The results presented in Figure 5 demonstrates that the proposed system was still able to perform the detection of moving edges reliably. This is largely attributed to the fact that there are inherent redundancies in the network, as each node is interconnected to 6 neighbouring nodes. This is particularly important as the fabrication yield of memristive devices is relatively poor given that the memristors’ technological readiness is rather low. The scalability of memristive technology indeed opens the possibility of achieving reliability through redundancy. 

\begin{figure}[h]
\includegraphics[scale = 0.6]{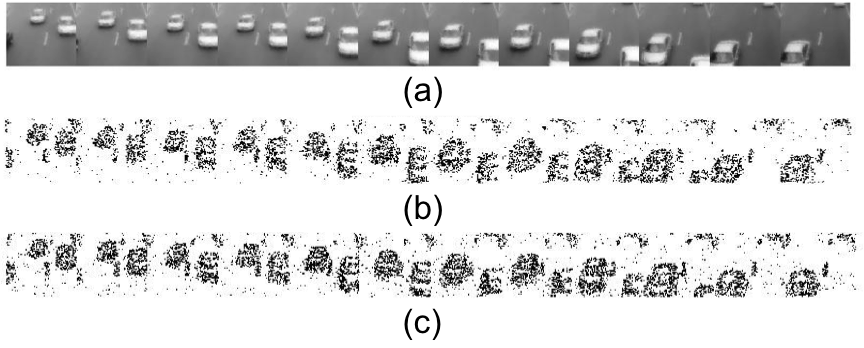}
\DeclareGraphicsExtensions{.png}
\caption{(a) Evolution of the smoothed input stimulus. (b) Moving edge detection with artificially simulated faults of 50\%. (c) Moving edge detection with artificially simulated faults of 80\%. }
\end{figure}

\section{\label{sec:level1} Double layered memristive network}
In the previous section, it was demonstrated that by employing the proposed thresholding scheme, memristive networks were able to perform moving edge detection, even with artificially simulated faults and at varying luminance levels. At the same time, this implementation is able to achieve local Gaussian filtering with a lower complexity as compared to other systems\cite{RefWorks:19} and has been shown to perform better than resistive grids for edge preserving smoothing \cite{RefWorks:79}. However, with this implementation, full optical flow computation is still not quite possible, as we have not yet obtained enough information from the sequence of moving images. The derivation of optical flow (in section II.C) shows that Equation (17) has to be solved in order to obtain the required optical flow. This requires both the spatial derivatives and the temporal derivatives at each pixel but our current implementation only provides us the spatial derivative (i.e. through the rate of change of memristance of memristive fuses). In this section, we present a 3D memristive grid that facillitates the computation of the temporal derivative.
\subsection{\label{sec:level2}Biomimetic OPL and IPL}
From the results in the previous section, we note that a second memristive layer is required to model the amacrine cells within the IPL of the retina that are responsible for generating the transient responses. Our model does not explicitly emulate the functioning of the neurons in the IPL, but rather the synaptic interconnections between them. In order to generate the transient responses, a delay of the input stimulus is required, and this is introduced by the delayed version of the input in the second layer.

\begin{figure}[h]
\includegraphics[scale = 0.65]{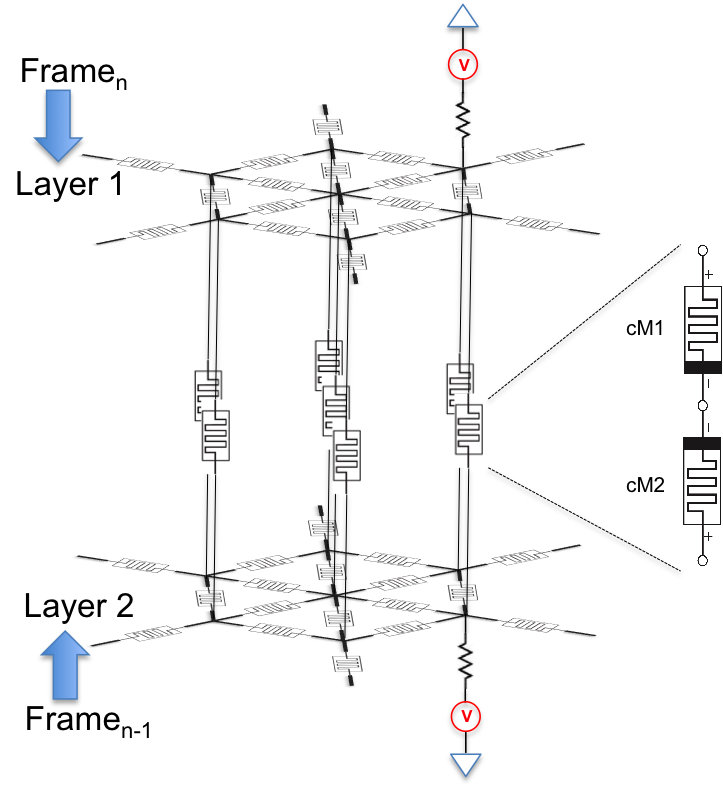}
\DeclareGraphicsExtensions{.png}
\caption{Illustration of a section of the proposed 3D hexagonal memristive network model. Each node corresponding to a specific pixel is connected in a hexagonal manner to 6 neighbouring nodes via memristive fuses. The network consists of two interconnected layers. The current signal is applied to Layer 1 while a delayed signal of 1 frame is applied to layer 2. Each pixel in the input frame is converted into a piecewise-linear voltage value, which is then applied through a resistor to the memristive grid. For clarity, only two of these biasing connections are shown.}
\end{figure}

The 3D network, part of which is illustrated in Figure 6, consists of 2 hexagonal memristive layers interconnected with memristive fuses at each node. The current signal is applied to layer 1, which is responsible for computing the edges from the sequence of frames. Whereas, in layer 2, a delayed version of each frame is used, enabling the computation of the temporal derivative at each node via the memristance changes in the interconnecting memristive fuses. 

Each connecting memristive fuse is made up of two individual memristors. We label the two memristor in each memristor fuse as Connecting Memristor 1 (cM1) and Connecting Memristor 2 (cM2), where cM1 is the memristor closer to layer 1 and cM2 is the one closer to layer 2. 

When an object in the input stimulus is completely stationary, it implies that the pixels corresponding to that object remains the same for the current frame n and the delayed frame (n-1). Consequently, the spatial voltage input (representing this object) to each layer will be identical and both ends of all connecting memristors are at equipotential. No current would thus flow and there will be no observable change in memristance. 

Conversely, when an object in the input stimulus moves, the pixels corresponding to that object will move. This means that the current frame n is slightly different from the delayed frame (n-1), reflecting this movement. Consider the case where a black box moves to the right by 2 pixels on a white background, as shown in Figure 7a. It can be observed that the rightmost and leftmost black edges of the box will be moving by 2 pixels to the right. Comparing with frame (n-1), the black box in frame n will be a shifted version by 2 pixels to the right. Hence, there will be a potential difference between the nodes corresponding to the new rightmost black edge (in frame n) and the old leftmost black edge (in frame (n-1)). At the new rightmost black edge, there will be a higher potential at layer 2 compared to layer 1. This causes a current flow through the connecting memristor fuse in from layer 2 to layer 1. Therefore, it is expected that the memristance cM2 will be increasing while the memristance of cM1 will be decreasing. Whereas for the old leftmost black edge, there will be a higher potential at layer 1 compared to layer 2. A current will flow from layer 1 to layer 2 causing the memristance of cM1 to increase and cM2 to decrease. 

Therefore, by monitoring the memristance changes in each memristor of each connecting memristor fuse, it is possible to determine the movement of the edges. For a dark object moving on a light background, decreasing cM1 and increasing cM2 corresponds to an appearing edge while increasing cM1 and an decrease of cM2 corresponds to a disappearing edge. This is similar to the on-center response in the retina. On the other hand, for a light object moving on a dark background, increasing cM1 and decreasing cM2 corresponds to an appearing edge while decreasing cM1 and increasing cM2 corresponds to a disappearing edge. This is similar to the off-center response of the retina. Essentially, the response of these connecting memristors captures the response of the transient amacrine cells in the retina.

With a delayed version of the input into the second layer of memristive network, it is possible to derive the temporal derivative between consecutive frames. The temporal derivative between corresponding pixels in consecutive frames is inherently computed as changes in memristance of the memristors within the connecting memristive fuse. By combining the temporal derivative information with the spatial derivative information obtained from the hexagonal memristive network, it will be possible to separate moving objects from stationary objects. Essentially the hexagonal grid is performing computation of the sustained response while the connecting memristors are responsible for computing the transient response. 

\begin{figure*}
\includegraphics[width=\textwidth]{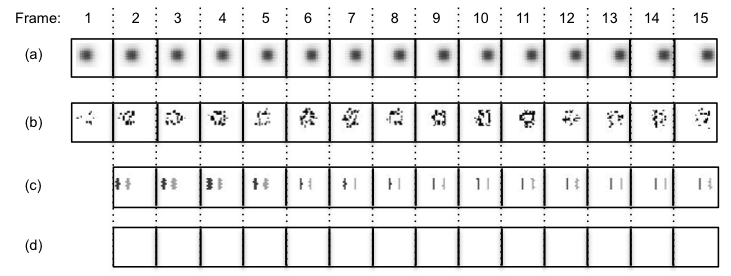}
\DeclareGraphicsExtensions{.png}
\caption{\label{fig:blackbox}(a) Evolution of input stimulus: A black box moving to the right at a rate of 2 pixels per frame from frame1 to frame 5, and 1 pixels per frame from frame6 to frame 15. (b) Simulation result of moving edge detection. (c) OFF-center transient detection results from monitoring the connecting memristive fuses. (d) ON-center transient detection results from monitoring the connecting memristive fuses. }
\end{figure*}

\subsection{\label{sec:level2}Emulating Transient Detection}
\subsubsection{Simple Inputs}
As an initial proof of how the abovementioned motion computation can be done with the double-layered memristive network, two simple simulations were performed. Each of the simulation input stimulus consists of a 1s, 15 fps, 17x30 pixels video clip. 

The first simulation, illustrated in Figure \ref{fig:blackbox}, involves a black box moving across a white background. From frame 1 to frame 5, the box is moving to the right at a rate of 2 pixels per frame and from frame 6 to frame 15, the box continues moving to the right at a slower rate of 1 pixel per frame. Since the moving object is dark compared to the surrounding, the on-center transient response (Figure 7(d)) is zero, which is expected. In the off-center transient response (Figure 7(c)), the light grey pixel corresponds to the appearing edge, while the dark pixel corresponds to disappearing edges. We see that the model is capable of  monitoring the movement of the edges, while the speed of this movement can be derived from the thickness of the transient edges. 

Our second example is similar to the first except for the fact that instead of a black box moving across a white background, a white box is now moving across a black background. In this case, the off-center transient response is zero and all the transient response is due to the on-center transient response. Here, the dark grey pixels correspond to an appearing edge while the light grey pixels correspond to a disappearing edge. The results are further illustrated in S5(see Supplementary Material\cite{Supplemental}).

\subsubsection{Complex Inputs}
The simulation was extended to a more complex input, which involves the first two frames of the traffic video stimulus shown in Figure 1a. Only twp frames were simulated as the simulation of more frames are rather computationally intensive. The frame size used in this simulation was 80x70pixel. This results in a circuit with 22400 nodes, 38602 memristive fuses, 11200 resistors and 11200 varying voltage sources. Even when running on a HP Z820 workstation with two 8-core Intel Xeon 2Ghz processor and 64 GB of RAM, the simulation environment (HSPICE) is incompetent in solving more extended cases.

\begin{figure}[ht]
\includegraphics[scale = 1.5]{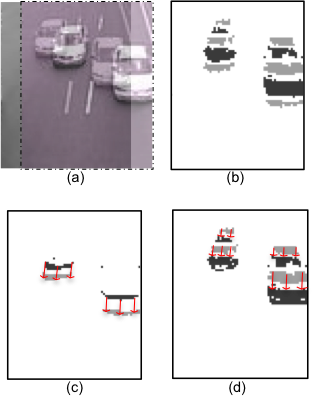}
\DeclareGraphicsExtensions{.png}
\caption{\label{fig:transient_complex} Simulation of 2 frames from a video of traffic on a highway. (a) Input stimulus consists of 2 frames. Frame 2 is overlaid on Frame 1, and for clarity, it is shifted slightly to the right and it has purple colour saturation (b) Transient detection (c) OFF-Center response (d) ON-center response}
\end{figure}
The input stimulus, shown in Figure  8(a) consists of 2 frames. Frame 2 is overlaid on Frame 1, and for clarity, is shifted slightly to the right and it has purple colour saturation. From Frame 1 to Frame 2, it can be observed that the 2 white cars have moved slightly forward down the road, while the black car has almost moved off the image. The simulation was performed for 1 frame period of 1/15 s.  By monitoring the memristance changes of the memristors within the connecting fuse, the results are shown on Figure 8(c) and (d) are obtained. From Figure 8(b), we do not consider the effects of the on-center or off-center response. The lighter grey dots correspond to decreasing cM1 and increasing cM2 while the darker grey dots corresponds to the increasing cM1 and decreasing cM2. This result shows the computation of transient response, where only moving objects were picked up. 

In order to accurately compute the direction and speed of the object, it is necessary to separate the movement of dark edges from the movement of light edges. Essentially, the on-center response and the off-center response have to be distinguished. This can be done by splitting the result in acquired in Figure 8(b) based on the intensity of the input stimulus to each pixel. The off-center response is shown in Figure 8(c), where only the transient response due to a moving dark object/edge is shown. The darker dots correspond to a disappearing dark edge while the lighter dots correspond to an appearing dark edge. Similarly, the on-center response, showing the transient response due to a moving light object/ edge is shown in Figure 8(d). The darker dots now correspond to an appearing light edge while the lighter dots correspond to a disappearing light edge.

\begin{figure}[t]
\includegraphics[scale = 1.1]{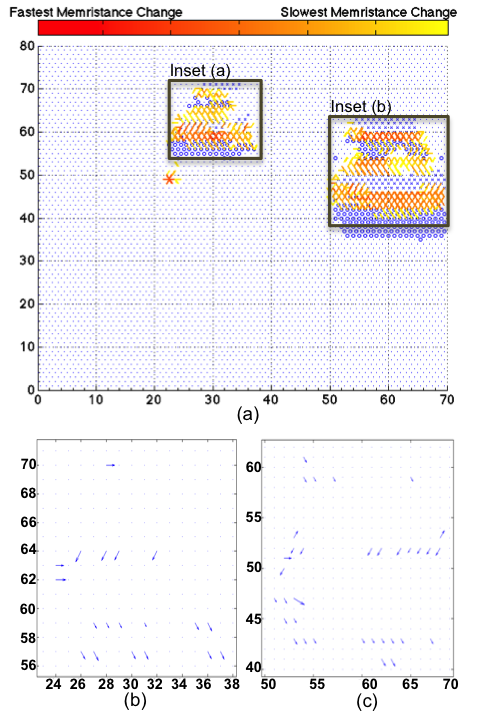}
\caption{\label{fig:DS_1} (a) Results illustrating the rate of memristance change of fuses within the 80x70pixel network. Each memristive fuse is plotted with a color corresponding to the rate at which it's memristance changes. (b) Illustrates the optical flow estimate from Inset (a). (c) Illustrates the optical flow estimate from inset (b).}
\end{figure}

\subsection{\label{sec:level2}Directional and Speed detection}
Given the output from the transient detection, it is possible to use a 1-layer hexagonal memristive grid to compute the direction and speed of the moving object. The results from the transient detection shown in Figure 8(c) and (d) can be used as the input stimuli. To demonstrate this, the memristive grid was biased based on the the off-center response from Figure 8(d). The darker pixels will correspond to a positive voltage source into the memristive network while the lighter pixels will correspond to a ground,. Figure 9(a) illustrates the 80x70pixel hexagonal memristive grid where the voltage sources and ground are marked by \textquotedblleft x\textquotedblright and \textquotedblleft o\textquotedblright respectively. The rate of memristance changes of all interconnecting fuses are monitored and they are drawn in Figure 9(a) with a value dependent colour variation. Since it is expected that the current will take the shortest path from the source to the ground, the rate of memristance changes of the fuses will provide an indication of the direction and flow of the current. At each of the moving edge pixel, an estimate of the optical flow can be computed by the vectorial sum of the rate of memristance change of the 6 interconnecting fuses. Figure 9(b) and 9(c) respectively shows the results of this computation for Inset (a) and Inset (b) of Figure 9(a). Our method provides local optical flow estimates at the brightness edges, and as illustrated in Figure 9(b) and 9(c), was able to correctly determine that both cars were moving downwards.

\section{\label{sec:level1}Conclusion}
In this paper, we presented a novel method for computing motion by leveraging on the scalability and non-linearity of memristors for emulating both the OPL and IPL of a vertebrate’s retina. The parallel processing nature of the proposed memristive network offers great potential in minimizing the computation and power requirements of conventional motion detection systems. A novel memristive thresholding scheme was introduced to allow the detection of moving edges. In addition, a double-layered memristive network was constructed to enable to detection of ON-center and OFF-center transient responses. Applying the transient detection results into a 1-D memristive network, it was then shown that it is possible to generate an estimate of the speed and direction of the moving object. This study serves as a good proof of concept that memristive networks have indeed great potential in performing motion computation effectively and efficiently.

\section{\label{sec:level1}Acknowledgement}
The authors wish to acknowledge the financial support of the CHIST-ERA ERAnet EPSRC EP/J00801X/1 and EP/K017829/1.


\bibliography{citations}

\providecommand{\noopsort}[1]{}\providecommand{\singleletter}[1]{#1}%
\begin{thebibliography}{57}%
\makeatletter
\providecommand \@ifxundefined [1]{%
 \@ifx{#1\undefined}
}%
\providecommand \@ifnum [1]{%
 \ifnum #1\expandafter \@firstoftwo
 \else \expandafter \@secondoftwo
 \fi
}%
\providecommand \@ifx [1]{%
 \ifx #1\expandafter \@firstoftwo
 \else \expandafter \@secondoftwo
 \fi
}%
\providecommand \natexlab [1]{#1}%
\providecommand \enquote  [1]{``#1''}%
\providecommand \bibnamefont  [1]{#1}%
\providecommand \bibfnamefont [1]{#1}%
\providecommand \citenamefont [1]{#1}%
\providecommand \href@noop [0]{\@secondoftwo}%
\providecommand \href [0]{\begingroup \@sanitize@url \@href}%
\providecommand \@href[1]{\@@startlink{#1}\@@href}%
\providecommand \@@href[1]{\endgroup#1\@@endlink}%
\providecommand \@sanitize@url [0]{\catcode `\\12\catcode `\$12\catcode
  `\&12\catcode `\#12\catcode `\^12\catcode `\_12\catcode `\%12\relax}%
\providecommand \@@startlink[1]{}%
\providecommand \@@endlink[0]{}%
\providecommand \url  [0]{\begingroup\@sanitize@url \@url }%
\providecommand \@url [1]{\endgroup\@href {#1}{\urlprefix }}%
\providecommand \urlprefix  [0]{URL }%
\providecommand \Eprint [0]{\href }%
\providecommand \doibase [0]{http://dx.doi.org/}%
\providecommand \selectlanguage [0]{\@gobble}%
\providecommand \bibinfo  [0]{\@secondoftwo}%
\providecommand \bibfield  [0]{\@secondoftwo}%
\providecommand \translation [1]{[#1]}%
\providecommand \BibitemOpen [0]{}%
\providecommand \bibitemStop [0]{}%
\providecommand \bibitemNoStop [0]{.\EOS\space}%
\providecommand \EOS [0]{\spacefactor3000\relax}%
\providecommand \BibitemShut  [1]{\csname bibitem#1\endcsname}%
\let\auto@bib@innerbib\@empty
\bibitem [{\citenamefont {Davies}(2004)}]{RefWorks:145}%
  \BibitemOpen
  \bibfield  {author} {\bibinfo {author} {\bibfnamefont {E.~R.}\ \bibnamefont
  {Davies}},\ }\href@noop {} {\emph {\bibinfo {title} {Machine Vision: Theory,
  Algorithms, Practicalities}}}\ (\bibinfo  {publisher} {Morgan Kaufmann
  Publishers Inc},\ \bibinfo {address} {San Francisco, CA, USA},\ \bibinfo
  {year} {2004})\BibitemShut {NoStop}%
\bibitem [{\citenamefont {Steger}\ \emph {et~al.}(2008)\citenamefont {Steger},
  \citenamefont {Ulrich},\ and\ \citenamefont {Wiedemann}}]{RefWorks:143}%
  \BibitemOpen
  \bibfield  {author} {\bibinfo {author} {\bibfnamefont {C.}~\bibnamefont
  {Steger}}, \bibinfo {author} {\bibfnamefont {M.}~\bibnamefont {Ulrich}}, \
  and\ \bibinfo {author} {\bibfnamefont {C.}~\bibnamefont {Wiedemann}},\
  }\href@noop {} {\emph {\bibinfo {title} {Machine Vision Algorithms and
  Applications}}}\ (\bibinfo {year} {2008})\BibitemShut {NoStop}%
\bibitem [{\citenamefont {Malamas}\ \emph {et~al.}(2003)\citenamefont
  {Malamas}, \citenamefont {Petrakis}, \citenamefont {Zervakis}, \citenamefont
  {Petit},\ and\ \citenamefont {didier Legat}}]{RefWorks:149}%
  \BibitemOpen
  \bibfield  {author} {\bibinfo {author} {\bibfnamefont {E.~N.}\ \bibnamefont
  {Malamas}}, \bibinfo {author} {\bibfnamefont {E.~G.~M.}\ \bibnamefont
  {Petrakis}}, \bibinfo {author} {\bibfnamefont {M.}~\bibnamefont {Zervakis}},
  \bibinfo {author} {\bibfnamefont {L.}~\bibnamefont {Petit}}, \ and\ \bibinfo
  {author} {\bibfnamefont {J.}~\bibnamefont {didier Legat}},\ }\href@noop {}
  {\bibfield  {journal} {\bibinfo  {journal} {Image and Vision Computing}\
  }\textbf {\bibinfo {volume} {21}},\ \bibinfo {pages} {171} (\bibinfo {year}
  {2003})}\BibitemShut {NoStop}%
\bibitem [{\citenamefont {Shafarenko}\ \emph {et~al.}(1997)\citenamefont
  {Shafarenko}, \citenamefont {Petrou},\ and\ \citenamefont
  {Kittler}}]{RefWorks:148}%
  \BibitemOpen
  \bibfield  {author} {\bibinfo {author} {\bibfnamefont {L.}~\bibnamefont
  {Shafarenko}}, \bibinfo {author} {\bibfnamefont {M.}~\bibnamefont {Petrou}},
  \ and\ \bibinfo {author} {\bibfnamefont {J.}~\bibnamefont {Kittler}},\ }\href
  {http://dx.doi.org/10.1109/83.641413} {\bibfield  {journal} {\bibinfo
  {journal} {Trans.Img.Proc.}\ }\textbf {\bibinfo {volume} {6}},\ \bibinfo
  {pages} {1530} (\bibinfo {year} {1997})}\BibitemShut {NoStop}%
\bibitem [{\citenamefont {Sanz}\ and\ \citenamefont
  {Petkovic}(1988)}]{RefWorks:146}%
  \BibitemOpen
  \bibfield  {author} {\bibinfo {author} {\bibfnamefont {J.~L.~C.}\
  \bibnamefont {Sanz}}\ and\ \bibinfo {author} {\bibfnamefont {D.}~\bibnamefont
  {Petkovic}},\ }\href@noop {} {\bibfield  {journal} {\bibinfo  {journal}
  {Pattern Analysis and Machine Intelligence, IEEE Transactions on}\ }\textbf
  {\bibinfo {volume} {10}},\ \bibinfo {pages} {830} (\bibinfo {year}
  {1988})}\BibitemShut {NoStop}%
\bibitem [{\citenamefont {Li}\ and\ \citenamefont {Lin}(1994)}]{RefWorks:147}%
  \BibitemOpen
  \bibfield  {author} {\bibinfo {author} {\bibfnamefont {H.}~\bibnamefont
  {Li}}\ and\ \bibinfo {author} {\bibfnamefont {J.~C.}\ \bibnamefont {Lin}},\
  }\href@noop {} {\bibfield  {journal} {\bibinfo  {journal} {Industry
  Applications, IEEE Transactions on}\ }\textbf {\bibinfo {volume} {30}},\
  \bibinfo {pages} {317} (\bibinfo {year} {1994})}\BibitemShut {NoStop}%
\bibitem [{\citenamefont {chii Liu}\ and\ \citenamefont
  {Usseglio-Viretta}(2000)}]{RefWorks:142}%
  \BibitemOpen
  \bibfield  {author} {\bibinfo {author} {\bibfnamefont {S.}~\bibnamefont {chii
  Liu}}\ and\ \bibinfo {author} {\bibfnamefont {A.}~\bibnamefont
  {Usseglio-Viretta}},\ }in\ \href@noop {} {\emph {\bibinfo {booktitle} {in 2nd
  Int. ICSC Symp. Neural Comput}}}\ (\bibinfo {year} {2000})\ pp.\ \bibinfo
  {pages} {23--26}\BibitemShut {NoStop}%
\bibitem [{\citenamefont {Huber}\ \emph {et~al.}(1999)\citenamefont {Huber},
  \citenamefont {Franz},\ and\ \citenamefont {BÃ¼lthoff}}]{RefWorks:144}%
  \BibitemOpen
  \bibfield  {author} {\bibinfo {author} {\bibfnamefont {S.~A.}\ \bibnamefont
  {Huber}}, \bibinfo {author} {\bibfnamefont {M.~O.}\ \bibnamefont {Franz}}, \
  and\ \bibinfo {author} {\bibfnamefont {H.~H.}\ \bibnamefont {BÃ¼lthoff}},\
  }\href@noop {} {\bibfield  {journal} {\bibinfo  {journal} {Robotics and
  Autonomous Systems}\ }\textbf {\bibinfo {volume} {29}},\ \bibinfo {pages}
  {227} (\bibinfo {year} {1999})}\BibitemShut {NoStop}%
\bibitem [{\citenamefont {Sotelo}\ \emph {et~al.}(2004)\citenamefont {Sotelo},
  \citenamefont {Rodriguez},\ and\ \citenamefont {Magdalena}}]{RefWorks:150}%
  \BibitemOpen
  \bibfield  {author} {\bibinfo {author} {\bibfnamefont {M.~A.}\ \bibnamefont
  {Sotelo}}, \bibinfo {author} {\bibfnamefont {F.~J.}\ \bibnamefont
  {Rodriguez}}, \ and\ \bibinfo {author} {\bibfnamefont {L.}~\bibnamefont
  {Magdalena}},\ }\href@noop {} {\bibfield  {journal} {\bibinfo  {journal}
  {Intelligent Transportation Systems, IEEE Transactions on}\ }\textbf
  {\bibinfo {volume} {5}},\ \bibinfo {pages} {69} (\bibinfo {year}
  {2004})}\BibitemShut {NoStop}%
\bibitem [{\citenamefont {Masaki}(1998)}]{RefWorks:151}%
  \BibitemOpen
  \bibfield  {author} {\bibinfo {author} {\bibfnamefont {I.}~\bibnamefont
  {Masaki}},\ }\href@noop {} {\bibfield  {journal} {\bibinfo  {journal}
  {Intelligent Systems and their Applications, IEEE}\ }\textbf {\bibinfo
  {volume} {13}},\ \bibinfo {pages} {24} (\bibinfo {year} {1998})}\BibitemShut
  {NoStop}%
\bibitem [{\citenamefont {Canals}\ \emph {et~al.}(2002)\citenamefont {Canals},
  \citenamefont {Roussel}, \citenamefont {Famechon},\ and\ \citenamefont
  {Treuillet}}]{RefWorks:152}%
  \BibitemOpen
  \bibfield  {author} {\bibinfo {author} {\bibfnamefont {R.}~\bibnamefont
  {Canals}}, \bibinfo {author} {\bibfnamefont {A.}~\bibnamefont {Roussel}},
  \bibinfo {author} {\bibfnamefont {J.~L.}\ \bibnamefont {Famechon}}, \ and\
  \bibinfo {author} {\bibfnamefont {S.}~\bibnamefont {Treuillet}},\ }\href@noop
  {} {\bibfield  {journal} {\bibinfo  {journal} {Industrial Electronics, IEEE
  Transactions on}\ }\textbf {\bibinfo {volume} {49}},\ \bibinfo {pages} {500}
  (\bibinfo {year} {2002})}\BibitemShut {NoStop}%
\bibitem [{\citenamefont {Saripalli}\ \emph {et~al.}(2003)\citenamefont
  {Saripalli}, \citenamefont {Montgomery},\ and\ \citenamefont
  {Sukhatme}}]{RefWorks:153}%
  \BibitemOpen
  \bibfield  {author} {\bibinfo {author} {\bibfnamefont {S.}~\bibnamefont
  {Saripalli}}, \bibinfo {author} {\bibfnamefont {J.~F.}\ \bibnamefont
  {Montgomery}}, \ and\ \bibinfo {author} {\bibfnamefont {G.~S.}\ \bibnamefont
  {Sukhatme}},\ }\href@noop {} {\bibfield  {journal} {\bibinfo  {journal}
  {Robotics and Automation, IEEE Transactions on}\ }\textbf {\bibinfo {volume}
  {19}},\ \bibinfo {pages} {371} (\bibinfo {year} {2003})}\BibitemShut
  {NoStop}%
\bibitem [{\citenamefont {Capson}\ \emph {et~al.}(1989)\citenamefont {Capson},
  \citenamefont {Maludzinski},\ and\ \citenamefont
  {Feuerstein}}]{RefWorks:154}%
  \BibitemOpen
  \bibfield  {author} {\bibinfo {author} {\bibfnamefont {D.~W.}\ \bibnamefont
  {Capson}}, \bibinfo {author} {\bibfnamefont {R.~A.}\ \bibnamefont
  {Maludzinski}}, \ and\ \bibinfo {author} {\bibfnamefont {I.~A.}\ \bibnamefont
  {Feuerstein}},\ }\href@noop {} {\bibfield  {journal} {\bibinfo  {journal}
  {Biomedical Engineering, IEEE Transactions on}\ }\textbf {\bibinfo {volume}
  {36}},\ \bibinfo {pages} {860} (\bibinfo {year} {1989})}\BibitemShut
  {NoStop}%
\bibitem [{\citenamefont {Ray}\ \emph {et~al.}(2002)\citenamefont {Ray},
  \citenamefont {Acton},\ and\ \citenamefont {Ley}}]{RefWorks:155}%
  \BibitemOpen
  \bibfield  {author} {\bibinfo {author} {\bibfnamefont {N.}~\bibnamefont
  {Ray}}, \bibinfo {author} {\bibfnamefont {S.~T.}\ \bibnamefont {Acton}}, \
  and\ \bibinfo {author} {\bibfnamefont {K.}~\bibnamefont {Ley}},\ }\href@noop
  {} {\bibfield  {journal} {\bibinfo  {journal} {Medical Imaging, IEEE
  Transactions on}\ }\textbf {\bibinfo {volume} {21}},\ \bibinfo {pages} {1222}
  (\bibinfo {year} {2002})}\BibitemShut {NoStop}%
\bibitem [{\citenamefont {Netravali}(1989)}]{RefWorks:30}%
  \BibitemOpen
  \bibfield  {author} {\bibinfo {author} {\bibfnamefont {A.~N.}\ \bibnamefont
  {Netravali}},\ }in\ \href@noop {} {\emph {\bibinfo {booktitle} {Electrical
  and Electronics Engineers in Israel, 1989. The Sixteenth Conference of}}}\
  (\bibinfo {year} {1989})\ pp.\ \bibinfo {pages} {0--18}\BibitemShut {NoStop}%
\bibitem [{\citenamefont {Ullman}(1981)}]{RefWorks:158}%
  \BibitemOpen
  \bibfield  {author} {\bibinfo {author} {\bibfnamefont {S.}~\bibnamefont
  {Ullman}},\ }\href@noop {} {\bibfield  {journal} {\bibinfo  {journal}
  {Computer}\ }\textbf {\bibinfo {volume} {14}},\ \bibinfo {pages} {57}
  (\bibinfo {year} {1981})}\BibitemShut {NoStop}%
\bibitem [{\citenamefont {Mead}(1990)}]{RefWorks:159}%
  \BibitemOpen
  \bibfield  {author} {\bibinfo {author} {\bibfnamefont {C.}~\bibnamefont
  {Mead}},\ }\href@noop {} {\bibfield  {journal} {\bibinfo  {journal}
  {Proceedings of the IEEE}\ }\textbf {\bibinfo {volume} {78}},\ \bibinfo
  {pages} {1629} (\bibinfo {year} {1990})}\BibitemShut {NoStop}%
\bibitem [{\citenamefont {Turel}\ \emph {et~al.}(2004)\citenamefont {Turel},
  \citenamefont {Lee}, \citenamefont {Ma},\ and\ \citenamefont
  {Likharev}}]{RefWorks:160}%
  \BibitemOpen
  \bibfield  {author} {\bibinfo {author} {\bibnamefont {Turel}}, \bibinfo
  {author} {\bibfnamefont {J.~H.}\ \bibnamefont {Lee}}, \bibinfo {author}
  {\bibfnamefont {X.}~\bibnamefont {Ma}}, \ and\ \bibinfo {author}
  {\bibfnamefont {K.~K.}\ \bibnamefont {Likharev}},\ }\href
  {http://dx.doi.org/10.1002/cta.v32:5} {\bibfield  {journal} {\bibinfo
  {journal} {Int.J.Circuit Theory Appl.}\ }\textbf {\bibinfo {volume} {32}},\
  \bibinfo {pages} {277} (\bibinfo {year} {2004})}\BibitemShut {NoStop}%
\bibitem [{\citenamefont {Bhushan}(2009)}]{RefWorks:161}%
  \BibitemOpen
  \bibfield  {author} {\bibinfo {author} {\bibfnamefont {B.}~\bibnamefont
  {Bhushan}},\ }\href@noop {} {\bibfield  {journal} {\bibinfo  {journal}
  {Philosophical Transactions of the Royal Society A: Mathematical, Physical
  and Engineering Sciences}\ }\textbf {\bibinfo {volume} {367}},\ \bibinfo
  {pages} {1445} (\bibinfo {year} {2009})}\BibitemShut {NoStop}%
\bibitem [{\citenamefont {Hutchinson}\ \emph {et~al.}(1988)\citenamefont
  {Hutchinson}, \citenamefont {Koch}, \citenamefont {Luo},\ and\ \citenamefont
  {Mead}}]{RefWorks:127}%
  \BibitemOpen
  \bibfield  {author} {\bibinfo {author} {\bibfnamefont {J.}~\bibnamefont
  {Hutchinson}}, \bibinfo {author} {\bibfnamefont {C.}~\bibnamefont {Koch}},
  \bibinfo {author} {\bibfnamefont {J.}~\bibnamefont {Luo}}, \ and\ \bibinfo
  {author} {\bibfnamefont {C.}~\bibnamefont {Mead}},\ }\href@noop {} {\bibfield
   {journal} {\bibinfo  {journal} {Computer}\ }\textbf {\bibinfo {volume}
  {21}},\ \bibinfo {pages} {52} (\bibinfo {year} {1988})}\BibitemShut {NoStop}%
\bibitem [{\citenamefont {Mead}(1989)}]{RefWorks:121}%
  \BibitemOpen
  \bibfield  {author} {\bibinfo {author} {\bibfnamefont {C.}~\bibnamefont
  {Mead}},\ }\href@noop {} {\emph {\bibinfo {title} {Analog VLSI and neural
  systems}}}\ (\bibinfo  {publisher} {Addison-Wesley Longman Publishing Co.,
  Inc},\ \bibinfo {address} {Boston, MA, USA},\ \bibinfo {year}
  {1989})\BibitemShut {NoStop}%
\bibitem [{\citenamefont {Lichtsteiner}\ \emph {et~al.}(2008)\citenamefont
  {Lichtsteiner}, \citenamefont {Posch},\ and\ \citenamefont
  {Delbruck}}]{RefWorks:111}%
  \BibitemOpen
  \bibfield  {author} {\bibinfo {author} {\bibfnamefont {P.}~\bibnamefont
  {Lichtsteiner}}, \bibinfo {author} {\bibfnamefont {C.}~\bibnamefont {Posch}},
  \ and\ \bibinfo {author} {\bibfnamefont {T.}~\bibnamefont {Delbruck}},\
  }\href@noop {} {\bibfield  {journal} {\bibinfo  {journal} {Solid-State
  Circuits, IEEE Journal of}\ }\textbf {\bibinfo {volume} {43}},\ \bibinfo
  {pages} {566} (\bibinfo {year} {2008})}\BibitemShut {NoStop}%
\bibitem [{\citenamefont {Boahen}\ and\ \citenamefont
  {Bioengineering}(2002)}]{RefWorks:130}%
  \BibitemOpen
  \bibfield  {author} {\bibinfo {author} {\bibfnamefont {K.}~\bibnamefont
  {Boahen}}\ and\ \bibinfo {author} {\bibfnamefont {P.}~\bibnamefont
  {Bioengineering}},\ }in\ \href@noop {} {\emph {\bibinfo {booktitle}
  {INCREASING, ON, DECREASING, and OFF visual signals,â€ Analog Integrated
  Circuits and Signal Processing}}}\ (\bibinfo  {publisher} {Press},\ \bibinfo
  {year} {2002})\ pp.\ \bibinfo {pages} {121--135}\BibitemShut {NoStop}%
\bibitem [{\citenamefont {Baccus}\ \emph {et~al.}(2008)\citenamefont {Baccus},
  \citenamefont {Olveczky}, \citenamefont {Manu},\ and\ \citenamefont
  {Meister}}]{RefWorks:35}%
  \BibitemOpen
  \bibfield  {author} {\bibinfo {author} {\bibfnamefont {S.~A.}\ \bibnamefont
  {Baccus}}, \bibinfo {author} {\bibfnamefont {B.~P.}\ \bibnamefont
  {Olveczky}}, \bibinfo {author} {\bibfnamefont {M.}~\bibnamefont {Manu}}, \
  and\ \bibinfo {author} {\bibfnamefont {M.}~\bibnamefont {Meister}},\
  }\href@noop {} {\bibfield  {journal} {\bibinfo  {journal} {The Journal of
  neuroscience : the official journal of the Society for Neuroscience}\
  }\textbf {\bibinfo {volume} {28}},\ \bibinfo {pages} {6807} (\bibinfo {year}
  {2008})},\ \bibinfo {note} {jID: 8102140; ppublish}\BibitemShut {NoStop}%
\bibitem [{\citenamefont {Benson}\ and\ \citenamefont
  {Delbruck}(1991)}]{RefWorks:7}%
  \BibitemOpen
  \bibfield  {author} {\bibinfo {author} {\bibfnamefont {R.~G.}\ \bibnamefont
  {Benson}}\ and\ \bibinfo {author} {\bibfnamefont {T.}~\bibnamefont
  {Delbruck}},\ }in\ \href@noop {} {\emph {\bibinfo {booktitle} {Advances in
  Neural Information Processing Systems 4}}}\ (\bibinfo  {publisher} {Morgan
  Kaufmann},\ \bibinfo {year} {1991})\ pp.\ \bibinfo {pages}
  {756--763}\BibitemShut {NoStop}%
\bibitem [{\citenamefont {Borst}(2011)}]{RefWorks:49}%
  \BibitemOpen
  \bibfield  {author} {\bibinfo {author} {\bibfnamefont {A.}~\bibnamefont
  {Borst}},\ }\href@noop {} {\bibfield  {journal} {\bibinfo  {journal} {Current
  biology : CB}\ }\textbf {\bibinfo {volume} {21}},\ \bibinfo {pages} {R990}
  (\bibinfo {year} {2011})},\ \bibinfo {note} {cI: Copyright (c) 2011; JID:
  9107782; CON: Curr Biol. 2011 Dec 20;21(24):2077-84. PMID: 22137471;
  ppublish}\BibitemShut {NoStop}%
\bibitem [{\citenamefont {Chiu}\ and\ \citenamefont {Wu}(1997)}]{RefWorks:47}%
  \BibitemOpen
  \bibfield  {author} {\bibinfo {author} {\bibfnamefont {C.-F.}\ \bibnamefont
  {Chiu}}\ and\ \bibinfo {author} {\bibfnamefont {C.-Y.}\ \bibnamefont {Wu}},\
  }in\ \href@noop {} {\emph {\bibinfo {booktitle} {Circuits and Systems, 1997.
  ISCAS '97., Proceedings of 1997 IEEE International Symposium on}}},\
  Vol.~\bibinfo {volume} {1}\ (\bibinfo {year} {1997})\ pp.\ \bibinfo {pages}
  {717--720 vol.1}\BibitemShut {NoStop}%
\bibitem [{\citenamefont {Delbruck}(1993)}]{RefWorks:56}%
  \BibitemOpen
  \bibfield  {author} {\bibinfo {author} {\bibfnamefont {T.}~\bibnamefont
  {Delbruck}},\ }\href@noop {} {\bibfield  {journal} {\bibinfo  {journal}
  {Neural Networks, IEEE Transactions on}\ }\textbf {\bibinfo {volume} {4}},\
  \bibinfo {pages} {529} (\bibinfo {year} {1993})}\BibitemShut {NoStop}%
\bibitem [{\citenamefont {Eeckman}\ \emph {et~al.}(1989)\citenamefont
  {Eeckman}, \citenamefont {Colvin},\ and\ \citenamefont
  {Axelrod}}]{RefWorks:39}%
  \BibitemOpen
  \bibfield  {author} {\bibinfo {author} {\bibfnamefont {F.~H.}\ \bibnamefont
  {Eeckman}}, \bibinfo {author} {\bibfnamefont {M.~E.}\ \bibnamefont {Colvin}},
  \ and\ \bibinfo {author} {\bibfnamefont {T.~S.}\ \bibnamefont {Axelrod}},\
  }in\ \href@noop {} {\emph {\bibinfo {booktitle} {Neural Networks, 1989.
  IJCNN., International Joint Conference on}}}\ (\bibinfo {year} {1989})\ pp.\
  \bibinfo {pages} {247--249 vol.2}\BibitemShut {NoStop}%
\bibitem [{\citenamefont {Kameda}\ and\ \citenamefont
  {Yagi}(2003)}]{RefWorks:44}%
  \BibitemOpen
  \bibfield  {author} {\bibinfo {author} {\bibfnamefont {S.}~\bibnamefont
  {Kameda}}\ and\ \bibinfo {author} {\bibfnamefont {T.}~\bibnamefont {Yagi}},\
  }in\ \href@noop {} {\emph {\bibinfo {booktitle} {Circuits and Systems, 2003.
  ISCAS '03. Proceedings of the 2003 International Symposium on}}},\
  Vol.~\bibinfo {volume} {4}\ (\bibinfo {year} {2003})\ pp.\ \bibinfo {pages}
  {IV--792--IV--795 vol.4}\BibitemShut {NoStop}%
\bibitem [{\citenamefont {Kameda}\ and\ \citenamefont
  {Yagi}(2002)}]{RefWorks:46}%
  \BibitemOpen
  \bibfield  {author} {\bibinfo {author} {\bibfnamefont {S.}~\bibnamefont
  {Kameda}}\ and\ \bibinfo {author} {\bibfnamefont {T.}~\bibnamefont {Yagi}},\
  }in\ \href@noop {} {\emph {\bibinfo {booktitle} {SICE 2002. Proceedings of
  the 41st SICE Annual Conference}}},\ Vol.~\bibinfo {volume} {3}\ (\bibinfo
  {year} {2002})\ pp.\ \bibinfo {pages} {1853--1858 vol.3}\BibitemShut
  {NoStop}%
\bibitem [{\citenamefont {Meitzler}\ \emph {et~al.}(1995)\citenamefont
  {Meitzler}, \citenamefont {Strohbehn},\ and\ \citenamefont
  {Andreou}}]{RefWorks:40}%
  \BibitemOpen
  \bibfield  {author} {\bibinfo {author} {\bibfnamefont {R.~C.}\ \bibnamefont
  {Meitzler}}, \bibinfo {author} {\bibfnamefont {K.}~\bibnamefont {Strohbehn}},
  \ and\ \bibinfo {author} {\bibfnamefont {A.~G.}\ \bibnamefont {Andreou}},\
  }in\ \href@noop {} {\emph {\bibinfo {booktitle} {Circuits and Systems, 1995.
  ISCAS '95., 1995 IEEE International Symposium on}}},\ Vol.~\bibinfo {volume}
  {3}\ (\bibinfo {year} {1995})\ pp.\ \bibinfo {pages} {2096--2099
  vol.3}\BibitemShut {NoStop}%
\bibitem [{\citenamefont {Mhani}\ \emph {et~al.}(1995)\citenamefont {Mhani},
  \citenamefont {Bouvier},\ and\ \citenamefont {Herault}}]{RefWorks:42}%
  \BibitemOpen
  \bibfield  {author} {\bibinfo {author} {\bibfnamefont {A.}~\bibnamefont
  {Mhani}}, \bibinfo {author} {\bibfnamefont {G.}~\bibnamefont {Bouvier}}, \
  and\ \bibinfo {author} {\bibfnamefont {J.}~\bibnamefont {Herault}},\ }in\
  \href@noop {} {\emph {\bibinfo {booktitle} {Solid-State Circuits Conference,
  1995. ESSCIRC '95. Twenty-first European}}}\ (\bibinfo {year} {1995})\ pp.\
  \bibinfo {pages} {326--329}\BibitemShut {NoStop}%
\bibitem [{\citenamefont {Yang}\ \emph {et~al.}(2011)\citenamefont {Yang},
  \citenamefont {Lin}, \citenamefont {Chiueh},\ and\ \citenamefont
  {Wu}}]{RefWorks:41}%
  \BibitemOpen
  \bibfield  {author} {\bibinfo {author} {\bibfnamefont {W.-C.}\ \bibnamefont
  {Yang}}, \bibinfo {author} {\bibfnamefont {L.-J.}\ \bibnamefont {Lin}},
  \bibinfo {author} {\bibfnamefont {H.}~\bibnamefont {Chiueh}}, \ and\ \bibinfo
  {author} {\bibfnamefont {C.-Y.}\ \bibnamefont {Wu}},\ }\href@noop {}
  {\bibfield  {journal} {\bibinfo  {journal} {Sensors Journal, IEEE}\ }\textbf
  {\bibinfo {volume} {11}},\ \bibinfo {pages} {3341} (\bibinfo {year}
  {2011})}\BibitemShut {NoStop}%
\bibitem [{\citenamefont {Mahowald}(1994)}]{RefWorks:162}%
  \BibitemOpen
  \bibfield  {author} {\bibinfo {author} {\bibfnamefont {M.}~\bibnamefont
  {Mahowald}},\ }\href@noop {} {\emph {\bibinfo {title} {An Analog VLSI System
  for Stereoscopic Vision}}}\ (\bibinfo  {publisher} {Kluwer Academic
  Publishers},\ \bibinfo {address} {Norwell, MA, USA},\ \bibinfo {year}
  {1994})\BibitemShut {NoStop}%
\bibitem [{\citenamefont {Williams}(2008)}]{RefWorks:137}%
  \BibitemOpen
  \bibfield  {author} {\bibinfo {author} {\bibfnamefont {R.}~\bibnamefont
  {Williams}},\ }\href@noop {} {\bibfield  {journal} {\bibinfo  {journal}
  {Spectrum, IEEE}\ }\textbf {\bibinfo {volume} {45}},\ \bibinfo {pages} {28}
  (\bibinfo {year} {2008})}\BibitemShut {NoStop}%
\bibitem [{\citenamefont {Ebong}\ and\ \citenamefont
  {Mazumder}(2012)}]{RefWorks:23}%
  \BibitemOpen
  \bibfield  {author} {\bibinfo {author} {\bibfnamefont {I.~E.}\ \bibnamefont
  {Ebong}}\ and\ \bibinfo {author} {\bibfnamefont {P.}~\bibnamefont
  {Mazumder}},\ }\href@noop {} {\bibfield  {journal} {\bibinfo  {journal}
  {Proceedings of the IEEE}\ }\textbf {\bibinfo {volume} {100}},\ \bibinfo
  {pages} {2050} (\bibinfo {year} {2012})}\BibitemShut {NoStop}%
\bibitem [{\citenamefont {Linares-Barranco}\ \emph {et~al.}(2011)\citenamefont
  {Linares-Barranco}, \citenamefont {Serrano-Gotarredona}, \citenamefont
  {CamuÃ±as-Mesa}, \citenamefont {Perez-Carrasco}, \citenamefont
  {ZamarreÃ±o-Ramos},\ and\ \citenamefont {Masquelier}}]{RefWorks:27}%
  \BibitemOpen
  \bibfield  {author} {\bibinfo {author} {\bibfnamefont {B.}~\bibnamefont
  {Linares-Barranco}}, \bibinfo {author} {\bibfnamefont {T.}~\bibnamefont
  {Serrano-Gotarredona}}, \bibinfo {author} {\bibfnamefont {L.~A.}\
  \bibnamefont {CamuÃ±as-Mesa}}, \bibinfo {author} {\bibfnamefont {J.~A.}\
  \bibnamefont {Perez-Carrasco}}, \bibinfo {author} {\bibfnamefont
  {C.}~\bibnamefont {ZamarreÃ±o-Ramos}}, \ and\ \bibinfo {author}
  {\bibfnamefont {T.}~\bibnamefont {Masquelier}},\ }\href
  {http://www.frontiersin.org/neuromorphic_engineering/10.3389/fnins.2011.00026/abstract}
  {\bibfield  {journal} {\bibinfo  {journal} {Frontiers in Neuroscience}\
  }\textbf {\bibinfo {volume} {5}} (\bibinfo {year} {2011})}\BibitemShut
  {NoStop}%
\bibitem [{\citenamefont {Serrano-Gotarredona}\ \emph
  {et~al.}(2013)\citenamefont {Serrano-Gotarredona}, \citenamefont
  {Masquelier}, \citenamefont {Prodromakis}, \citenamefont {Indiveri},\ and\
  \citenamefont {Linares-Barranco}}]{RefWorks:163}%
  \BibitemOpen
  \bibfield  {author} {\bibinfo {author} {\bibfnamefont {T.}~\bibnamefont
  {Serrano-Gotarredona}}, \bibinfo {author} {\bibfnamefont {T.}~\bibnamefont
  {Masquelier}}, \bibinfo {author} {\bibfnamefont {T.}~\bibnamefont
  {Prodromakis}}, \bibinfo {author} {\bibfnamefont {G.}~\bibnamefont
  {Indiveri}}, \ and\ \bibinfo {author} {\bibfnamefont {B.}~\bibnamefont
  {Linares-Barranco}},\ }\href
  {http://www.frontiersin.org/neuroscience/10.3389/fnins.2013.00002/abstract}
  {\bibfield  {journal} {\bibinfo  {journal} {Frontiers in Neuroscience}\
  }\textbf {\bibinfo {volume} {7}} (\bibinfo {year} {2013})}\BibitemShut
  {NoStop}%
\bibitem [{\citenamefont {Gelencs\'er}\ \emph {et~al.}(2012)\citenamefont
  {Gelencs\'er}, \citenamefont {Prodromakis}, \citenamefont {Toumazou},\ and\
  \citenamefont {Roska}}]{RefWorks:19}%
  \BibitemOpen
  \bibfield  {author} {\bibinfo {author} {\bibfnamefont {A.}~\bibnamefont
  {Gelencs\'er}}, \bibinfo {author} {\bibfnamefont {T.}~\bibnamefont
  {Prodromakis}}, \bibinfo {author} {\bibfnamefont {C.}~\bibnamefont
  {Toumazou}}, \ and\ \bibinfo {author} {\bibfnamefont {T.}~\bibnamefont
  {Roska}},\ }\href {\doibase 10.1103/PhysRevE.85.041918} {\bibfield  {journal}
  {\bibinfo  {journal} {Phys. Rev. E}\ }\textbf {\bibinfo {volume} {85}},\
  \bibinfo {pages} {041918} (\bibinfo {year} {2012})}\BibitemShut {NoStop}%
\bibitem [{\citenamefont {Wandell}(1955)}]{RefWorks:82}%
  \BibitemOpen
  \bibfield  {author} {\bibinfo {author} {\bibfnamefont {B.~A.}\ \bibnamefont
  {Wandell}},\ }\href@noop {} {\emph {\bibinfo {title} {Foundation of
  Vision}}},\ \bibinfo {edition} {1st}\ ed.\ (\bibinfo  {publisher} {Sinauer
  Associates, Inc},\ \bibinfo {address} {United States of America},\ \bibinfo
  {year} {1955})\BibitemShut {NoStop}%
\bibitem [{\citenamefont {Dowling}(1987)}]{RefWorks:135}%
  \BibitemOpen
  \bibfield  {author} {\bibinfo {author} {\bibfnamefont {J.~E.}\ \bibnamefont
  {Dowling}},\ }\href@noop {} {\emph {\bibinfo {title} {The Retina - An
  Approachable Part of The Brain}}}\ (\bibinfo  {publisher} {The Belknap press
  of Harvard university press},\ \bibinfo {address} {USA},\ \bibinfo {year}
  {1987})\BibitemShut {NoStop}%
\bibitem [{\citenamefont {Regan}(2000)}]{RefWorks:136}%
  \BibitemOpen
  \bibfield  {author} {\bibinfo {author} {\bibfnamefont {D.}~\bibnamefont
  {Regan}},\ }\href@noop {} {\emph {\bibinfo {title} {Human Perception of
  Objects}}}\ (\bibinfo  {publisher} {Sinauer Associates, Inc, Publishers},\
  \bibinfo {address} {USA},\ \bibinfo {year} {2000})\BibitemShut {NoStop}%
\bibitem [{\citenamefont {Kolb}(2003)}]{RefWorks:3}%
  \BibitemOpen
  \bibfield  {author} {\bibinfo {author} {\bibfnamefont {H.}~\bibnamefont
  {Kolb}},\ }\href@noop {} {\bibfield  {journal} {\bibinfo  {journal} {American
  Scientist}\ }\textbf {\bibinfo {volume} {91}},\ \bibinfo {pages} {28}
  (\bibinfo {year} {2003})}\BibitemShut {NoStop}%
\bibitem [{\citenamefont {Lee}\ \emph {et~al.}(2001)\citenamefont {Lee},
  \citenamefont {Chae}, \citenamefont {Kim}, \citenamefont {Kim},\ and\
  \citenamefont {Cho}}]{RefWorks:43}%
  \BibitemOpen
  \bibfield  {author} {\bibinfo {author} {\bibfnamefont {J.~W.}\ \bibnamefont
  {Lee}}, \bibinfo {author} {\bibfnamefont {S.~P.}\ \bibnamefont {Chae}},
  \bibinfo {author} {\bibfnamefont {M.~N.}\ \bibnamefont {Kim}}, \bibinfo
  {author} {\bibfnamefont {S.~Y.}\ \bibnamefont {Kim}}, \ and\ \bibinfo
  {author} {\bibfnamefont {J.~H.}\ \bibnamefont {Cho}},\ }in\ \href@noop {}
  {\emph {\bibinfo {booktitle} {Industrial Electronics, 2001. Proceedings. ISIE
  2001. IEEE International Symposium on}}},\ Vol.~\bibinfo {volume} {1}\
  (\bibinfo {year} {2001})\ pp.\ \bibinfo {pages} {106--109 vol.1}\BibitemShut
  {NoStop}%
\bibitem [{\citenamefont {F.Y.}(2008)}]{RefWorks:97}%
  \BibitemOpen
  \bibfield  {author} {\bibinfo {author} {\bibfnamefont {W.}~\bibnamefont
  {F.Y.}},\ }\href {http://adsabs.harvard.edu/abs/2008arXiv0808.0286W}
  {\bibfield  {journal} {\bibinfo  {journal} {ArXiv e-prints}\ } (\bibinfo
  {year} {2008})},\ \bibinfo {note} {0808.0286; Provided by the SAO/NASA
  Astrophysics Data System}\BibitemShut {NoStop}%
\bibitem [{\citenamefont {Chua}\ and\ \citenamefont
  {Kang}(1976)}]{RefWorks:110}%
  \BibitemOpen
  \bibfield  {author} {\bibinfo {author} {\bibfnamefont {L.}~\bibnamefont
  {Chua}}\ and\ \bibinfo {author} {\bibfnamefont {S.~M.}\ \bibnamefont
  {Kang}},\ }\href@noop {} {\bibfield  {journal} {\bibinfo  {journal}
  {Proceedings of the IEEE}\ }\textbf {\bibinfo {volume} {64}},\ \bibinfo
  {pages} {209} (\bibinfo {year} {1976})}\BibitemShut {NoStop}%
\bibitem [{\citenamefont {Chua}(1971)}]{Chua1971}%
  \BibitemOpen
  \bibfield  {author} {\bibinfo {author} {\bibfnamefont {L.}~\bibnamefont
  {Chua}},\ }\href {\doibase 10.1109/TCT.1971.1083337} {\bibfield  {journal}
  {\bibinfo  {journal} {Circuit Theory, IEEE Transactions on}\ }\textbf
  {\bibinfo {volume} {18}},\ \bibinfo {pages} {507} (\bibinfo {year}
  {1971})}\BibitemShut {NoStop}%
\bibitem [{\citenamefont {Strukov}\ \emph {et~al.}(2008)\citenamefont
  {Strukov}, \citenamefont {Snider}, \citenamefont {Stewart},\ and\
  \citenamefont {Williams}}]{RefWorks:96}%
  \BibitemOpen
  \bibfield  {author} {\bibinfo {author} {\bibfnamefont {D.~B.}\ \bibnamefont
  {Strukov}}, \bibinfo {author} {\bibfnamefont {G.~S.}\ \bibnamefont {Snider}},
  \bibinfo {author} {\bibfnamefont {D.~R.}\ \bibnamefont {Stewart}}, \ and\
  \bibinfo {author} {\bibfnamefont {R.~S.}\ \bibnamefont {Williams}},\ }\href
  {\doibase 10.1038/nature06932} {\bibfield  {journal} {\bibinfo  {journal}
  {Nature}\ }\textbf {\bibinfo {volume} {453}},\ \bibinfo {pages} {80}
  (\bibinfo {year} {2008})}\BibitemShut {NoStop}%
\bibitem [{\citenamefont {Salaoru}\ \emph {et~al.}(2013)\citenamefont
  {Salaoru}, \citenamefont {Prodromakis}, \citenamefont {Khiat},\ and\
  \citenamefont {Toumazou}}]{salaoru2013}%
  \BibitemOpen
  \bibfield  {author} {\bibinfo {author} {\bibfnamefont {I.}~\bibnamefont
  {Salaoru}}, \bibinfo {author} {\bibfnamefont {T.}~\bibnamefont
  {Prodromakis}}, \bibinfo {author} {\bibfnamefont {A.}~\bibnamefont {Khiat}},
  \ and\ \bibinfo {author} {\bibfnamefont {C.}~\bibnamefont {Toumazou}},\
  }\href {\doibase 10.1063/1.4774089} {\bibfield  {journal} {\bibinfo
  {journal} {Applied Physics Letters}\ }\textbf {\bibinfo {volume} {102}},\
  \bibinfo {eid} {013506} (\bibinfo {year} {2013})}\BibitemShut {NoStop}%
\bibitem [{\citenamefont {Horn}\ and\ \citenamefont
  {Schunck}(1981)}]{RefWorks:83}%
  \BibitemOpen
  \bibfield  {author} {\bibinfo {author} {\bibfnamefont {B.~K.~P.}\
  \bibnamefont {Horn}}\ and\ \bibinfo {author} {\bibfnamefont {B.~G.}\
  \bibnamefont {Schunck}},\ }\href@noop {} {\enquote {\bibinfo {title}
  {Determining optical flow},}\ } (\bibinfo {year} {1981})\BibitemShut
  {NoStop}%
\bibitem [{\citenamefont {Barron}\ and\ \citenamefont
  {Thacker}(2005)}]{RefWorks:34}%
  \BibitemOpen
  \bibfield  {author} {\bibinfo {author} {\bibfnamefont {J.~L.}\ \bibnamefont
  {Barron}}\ and\ \bibinfo {author} {\bibfnamefont {N.~A.}\ \bibnamefont
  {Thacker}},\ }in\ \href@noop {} {\emph {\bibinfo {booktitle} {Tina Memo}}}\
  (\bibinfo {year} {2005})\ Chap.\ \bibinfo {chapter} {2004-012}\BibitemShut
  {NoStop}%
\bibitem [{\citenamefont {The~MathWorks}(2012)}]{Matlab_imageprocessing}%
  \BibitemOpen
  \bibfield  {author} {\bibinfo {author} {\bibfnamefont {I.}~\bibnamefont
  {The~MathWorks}},\ }\href {http://www.mathworks.co.uk/help/images/index.html}
  {\emph {\bibinfo {title} {MATLAB Image Processing Toolbox - User's Guide
  R2012b}}}\ (\bibinfo  {publisher} {The MathWorks, Inc},\ \bibinfo {address}
  {3 Apple Hill Drive, Natick, MA 01760-2098},\ \bibinfo {year}
  {2012})\BibitemShut {NoStop}%
\bibitem [{\citenamefont {Biolek}\ \emph {et~al.}(2009)\citenamefont {Biolek},
  \citenamefont {Biolek},\ and\ \citenamefont {Biolkova}}]{RefWorks:95}%
  \BibitemOpen
  \bibfield  {author} {\bibinfo {author} {\bibfnamefont {Z.}~\bibnamefont
  {Biolek}}, \bibinfo {author} {\bibfnamefont {D.}~\bibnamefont {Biolek}}, \
  and\ \bibinfo {author} {\bibfnamefont {V.}~\bibnamefont {Biolkova}},\
  }\href@noop {} {\bibfield  {journal} {\bibinfo  {journal} {Radioengineering}\
  }\textbf {\bibinfo {volume} {18}},\ \bibinfo {pages} {210} (\bibinfo {year}
  {2009})}\BibitemShut {NoStop}%
\bibitem [{\citenamefont {Prodromakis}\ \emph {et~al.}(2011)\citenamefont
  {Prodromakis}, \citenamefont {Peh}, \citenamefont {Papavassiliou},\ and\
  \citenamefont {Toumazou}}]{RefWorks:76}%
  \BibitemOpen
  \bibfield  {author} {\bibinfo {author} {\bibfnamefont {T.}~\bibnamefont
  {Prodromakis}}, \bibinfo {author} {\bibfnamefont {B.~P.}\ \bibnamefont
  {Peh}}, \bibinfo {author} {\bibfnamefont {C.}~\bibnamefont {Papavassiliou}},
  \ and\ \bibinfo {author} {\bibfnamefont {C.}~\bibnamefont {Toumazou}},\
  }\href@noop {} {\bibfield  {journal} {\bibinfo  {journal} {Electron Devices,
  IEEE Transactions on}\ }\textbf {\bibinfo {volume} {58}},\ \bibinfo {pages}
  {3099} (\bibinfo {year} {2011})}\BibitemShut {NoStop}%
\bibitem [{Sup()}]{Supplemental}%
  \BibitemOpen
  \href@noop {} {\enquote {\bibinfo {title} {See supplemental material at
  http://links.aps.org/supplemental/... for videos illustrating the time
  evolution of simulations presented in this paper.}}\ }\BibitemShut {NoStop}%
\bibitem [{\citenamefont {Jiang}\ and\ \citenamefont
  {Shi}(2009)}]{RefWorks:79}%
  \BibitemOpen
  \bibfield  {author} {\bibinfo {author} {\bibfnamefont {F.}~\bibnamefont
  {Jiang}}\ and\ \bibinfo {author} {\bibfnamefont {B.~E.}\ \bibnamefont
  {Shi}},\ }in\ \href@noop {} {\emph {\bibinfo {booktitle} {Circuit Theory and
  Design, 2009. ECCTD 2009. European Conference on}}}\ (\bibinfo {year}
  {2009})\ pp.\ \bibinfo {pages} {181--184}\BibitemShut {NoStop}%
\end{thebibliography}%

\end{document}